\documentclass[review]{elsarticle}

\usepackage{lineno,hyperref}

\modulolinenumbers[0]

\journal{Journal of Pattern Recognition}
\usepackage{graphicx}
\usepackage{subfigure}
\usepackage{amsmath,amssymb,amsthm}
\usepackage{multirow}

\usepackage{booktabs}
\usepackage{accents}
\usepackage{statex}
\usepackage{url}
\usepackage{color}
\usepackage{comment}

\usepackage[linesnumbered,ruled,vlined]{algorithm2e}
\SetKwInOut{Notation}{Notations}

\begin{document}

\begin{frontmatter}

\title{Learning Twofold Heterogeneous Multi-Task by Sharing Similar Convolution Kernel Pairs}

\author[mymainaddress1,mymainaddress2]{Quan Feng}
\author[mymainaddress1,mymainaddress2]{Songcan Chen\corref{mycorrespondingauthor}}
\cortext[mycorrespondingauthor]{Corresponding author}
\ead{s.chen@nuaa.edu.cn}

\address[mymainaddress1]{College of Computer Science $\&$ Technology, Nanjing University of Aeronautics $\&$ Astronautics, Nanjing, Jiangsu, 211106, China}
\address[mymainaddress2]{MIIT Key Laboratory of Pattern Analysis and Machine Intelligence, Nanjing University of Aeronautics $\&$ Astronautics, Nanjing, Jiangsu, 211106, China}

\begin{abstract}
Heterogeneous multi-task learning (HMTL) is an important topic in multi-task learning (MTL). Most existing HMTL methods usually solve either scenario where all tasks reside in the same input (feature) space yet unnecessarily the consistent output (label) space or scenario where their input (feature) spaces are heterogeneous while the output (label) space is consistent. However, to the best of our knowledge, there is limited study on twofold heterogeneous MTL (THMTL) scenario where the input and the output spaces are both inconsistent or heterogeneous. In order to handle this complicated scenario, in this paper, we design a simple and effective multi-task adaptive learning (MTAL) network to learn multiple tasks in such THMTL setting. Specifically, we explore and utilize the inherent relationship between tasks for knowledge sharing from similar convolution kernels in individual layers of the MTAL network. Then in order to realize the sharing, we weightedly aggregate any pair of convolutional kernels with their similarity greater than some threshold $\rho$, consequently, our model effectively performs cross-task learning while suppresses the intra-redundancy of the entire network. Finally, we conduct end-to-end training. Our experimental results demonstrate the effectiveness of our method in comparison with the state-of-the-art counterparts.
\end{abstract}

\begin{keyword}
heterogeneous tasks \sep multi-task learning \sep parameter sharing
\end{keyword}
\end{frontmatter}

\section{Introduction}
MTL aims to improve the generalization performance of individual tasks by learning multiple related tasks simultaneously. It has been successfully applied in computer vision \cite{kendall2018multi}, natural language processing \cite{dong2015multi}, speech recognition \cite{giri2015improving}, medical images processing \cite{Zhang2012Multi}, autonomous vehicles \cite{chen2018multi} and so on. Existing MTL methods assume that the inputs (feature) spaces and outputs (label) spaces of individual tasks are homogeneous (e.g., they have the same feature space and label space) \cite{long2017learning}, \cite{zhang2016embedding}, \cite{lin2015bilinear}. This assumption is only applicable in limited real world scenarios.

In practice, general HMTL methods also likewise assume that 1) the inputs (feature) spaces are homogeneous yet the outputs (label) spaces are inconsistent, for example, \cite{han2017heterogeneous} predicts the heterogeneous attributes of face images; \cite{li2014heterogeneous} evaluates poses of persons; 2) the input (feature) spaces are heterogeneous while the output (label) space is consistent, for example, \cite{luo2015large} uses the features of heterogeneous domains for image classification; \cite{8058002} learns heterogeneous domain metrics for text classification. However, there are more general scenarios in reality, where the input and output spaces are heterogeneous (i.e., twofold heterogeneity), thus posing a big challenge for HMTL. To the best of our knowledge, there has been almost no study yet to focus on the scenario.

In order to deal with such a complicated scenario, firstly, let us clarify the two unique issues involved that will affect the performance of the HMTL methods. The first is how to model the relationships among heterogeneous tasks as done for homogeneous tasks. Due to the existence of the heterogeneity among those tasks, there do more likely exist strong similarity, weak similarity, which in turn leads to positive, negative correlations \cite{zhao2019multiple}. Therefore, modeling such unknown relationships is important yet challenging. A recent work \cite{Schreiber2020EmergingRN} has shown that we can capture such relationship among tasks by embedding a shared unit into the multi-layer perceptron to reflect the specific information of tasks.
While by sharing the hyper-parameters, \cite{luvizon2020multi-task} designs a single pipeline HMTL network for heterogeneous pose evaluation and action recognition of humans.

Secondly, how to obtain useful shared information during the learning process. Since various information in heterogeneous tasks may have an implicit interrelationship with a single task or all tasks, thus undoubtedly affecting the performance of HMTL \cite{wu2020understanding}. Therefore, it is very important to obtain effective common information for individual tasks by the twofold heterogeneous multi-task (THMT) learning. Recent research work has designed a feature matching network to capture these shared information in heterogeneous tasks for HMTL \cite{liu2020multi}.

\begin{figure*}
\centering
\includegraphics[width=0.85\textwidth]{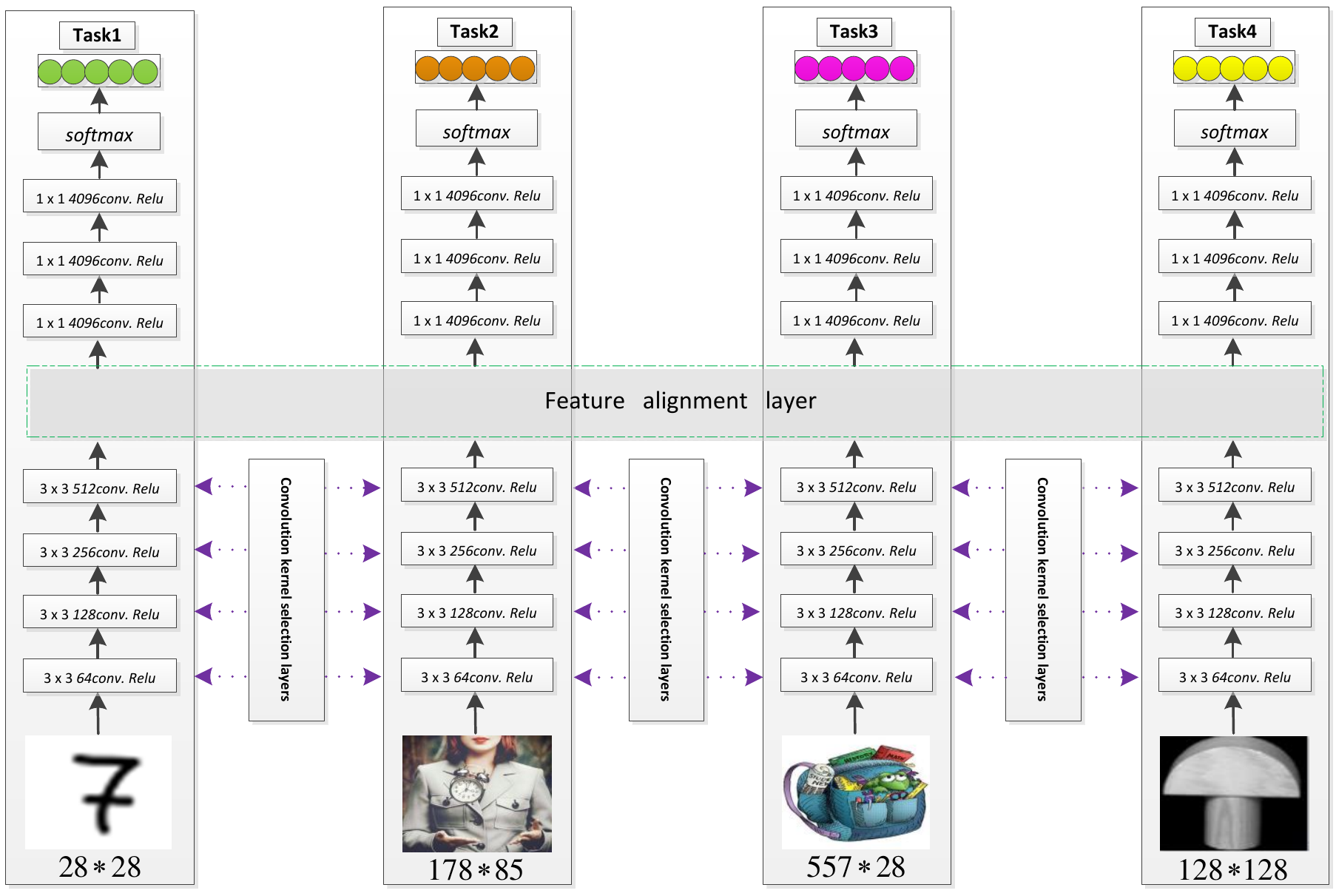}
\caption{Our proposed MTAL networks. The network is composed of four identical individual task networks, and the input/output task (feature/label) spaces
of each task network are heterogeneous, respectively. The black solid line mullion represents the convolution kernel selection layer. Green, orange, purple, and yellow solid circles indicate different output tasks.} \label{Fig:aa}
\label{fig1}
\end{figure*}

Even so, most previous HTML works relied on manually designing a shared layer for tasks including the latter layers until a branching layer. For example, \cite{dvornik2017blitznet:} uses a BlitzNet shared layer for object detection and semantic segmentation. \cite{DBLP:journals/corr/HuangFCY15} designs a dual-attribute-aware hierarchical network (DARN) for cross-domain image retrieval at fully connected layers. These constraints bring about comprehensive cross-task knowledge sharing. To overcome these, we designed a MTAL network. Different from existing methods, the designed network automatically selects similar convolution kernel pairs across-tasks to obtain common knowledge. Specifically, as shown in Fig \eqref{Fig:aa}: in the learning process, we automatically capture the inherent relationship between tasks by exploiting the similarity between convolution kernels. Next, we use a soft threshold to select the suitable convolution kernel pairs for aggregation to form a set of new kernel bank to learn individual tasks. Finally, we train the entire network in an end-to-end manner. Our experimental results demonstrate the effectiveness of our method in comparison with the state-of-the-art counter-parts.

In summary, our contribution can be summarized as  four folds below:

1) We propose a set of novel MTAL networks, which provides a new way of solving the information sharing problem in THMTL.

2) We design a soft threshold to select the suitable kernel pairs to prevent possible negative transfer caused by existing MTAL network learning.

3) We design a novel sharing strategy, which utilizes the similarity between convolution kernels in the same individual layers to formulate the intrinsic relationship between tasks and aggregate them to suppresses the intra-redundancy of the entire network.

4) We conduct experiments on eight public datasets and compare the state-of-the-art methods to validate the effectiveness of our method.

The rest of this article is organized as follows: in Section 2, we review the related methods of homogeneous and heterogeneous multi-task learnings respectively; in Section 3, we introduced the specific implementation of our method; in Section 4, we conduct experiments on the eight public benchmark datasets and compare the state-of-the-art methods to show the effectiveness of our method. Finally, Section 5 concludes this paper with future research directions. The code is available at {\color{blue} \emph {http://parnec.nuaa.edu.cn/3021/list.htm}}.

\section{Related Work}
At present, deep learning has achieved remarkable results in computer vision and MTL. Our work is also based on deep learning framework, thus in this section,
we mainly review deep learning-based homogeneous MTL and heterogeneous MTL in recent years, respectively.

\subsection{Homogeneous MTL}
Such methods generally assume that the input (feature) space and the output (label) space are homogeneous between tasks (i,e., the same feature space and label space). In such a scenario, existing methods share the same features between tasks, and they are applied in object detection, image classification, medical image segmentation, and so on. To date, there have had approaches proposed, we divide them into three categories according to the homogeneous MTL sharing architectures: hard sharing, soft sharing, and hybrid sharing.

\begin{figure}[ht]
\centering
\includegraphics[scale=0.53]{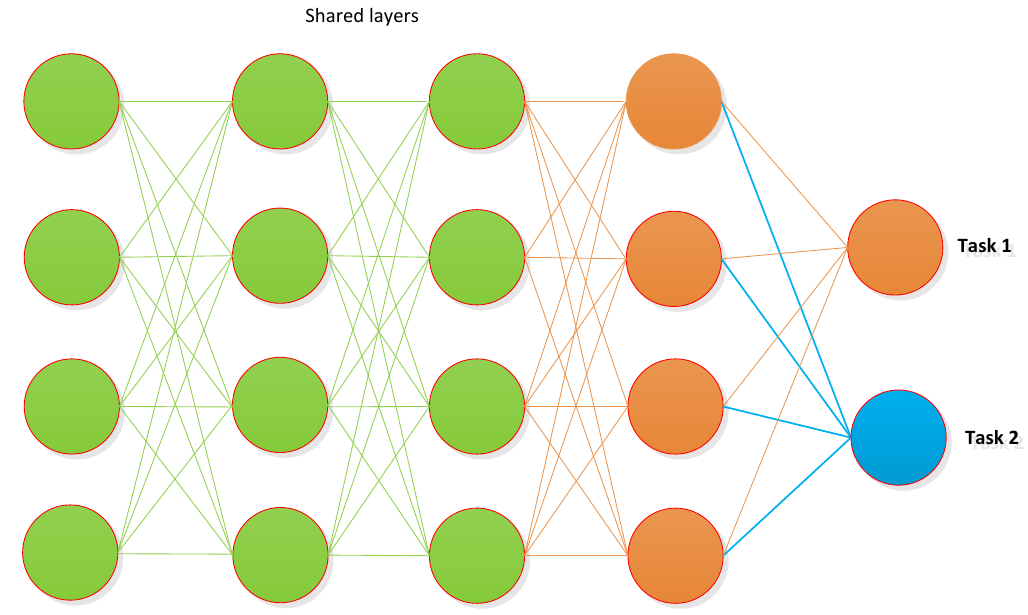}
\caption{Homogeneous MTL  based on hard sharing.} \label{Fig:bb}
\label{fig1}
\end{figure}

1) Hard sharing MTL. As shown in Fig.\eqref{Fig:bb}, in the architecture, all tasks share their knowledge in the same hidden space. For example, \cite{saeed2017personalized} uses a mutual representation of hard parameters shared for personalized stress recognition. \cite{Liu2020MultitaskLV} predicts the mortality of diverse rare diseases by initializing shared parameters in hidden space. The MTL methods of this kind can not only assist in effective learning between tasks but also minimize the risk of over fitting during the training process \cite{Baxter1997A}. However, it is difficult to handle loosely related tasks \cite{sun2020learning}.

\begin{figure}[ht]
\centering
\includegraphics[scale=0.53]{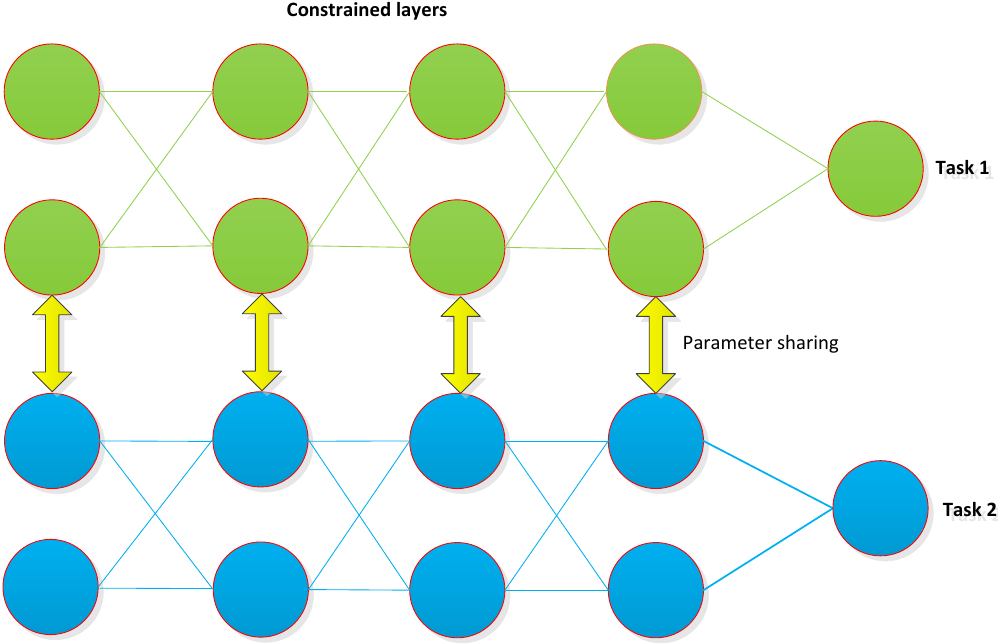}
\caption{Homogeneous MTL based on soft sharing.} \label{Fig:cc}
\label{fig1}
\end{figure}

2) Soft sharing MTL. As shown in Fig.\eqref{Fig:cc}, in its architecture, all task models and parameters are independent, and the distance between the
model parameters is regularized to obtain similar parameters for joint learning. For example, \cite{duong2015low} uses a neural network parser to share parameters between the source domain language and the target domain language for natural language processing. \cite{yang2016trace} utilizes tensor trace norms to regularize parameters in multiple networks for image recognition. \cite{NIPS2017_6757} learns multiple different tasks by jointly learning transferable features and multi-linear relationships between tasks and features in a fully connected layer. However, this method obviously relies on a predefined shared structure, and the model has poor generalization performance for the new tasks.

\begin{figure}[ht]
\centering
\includegraphics[scale=0.54]{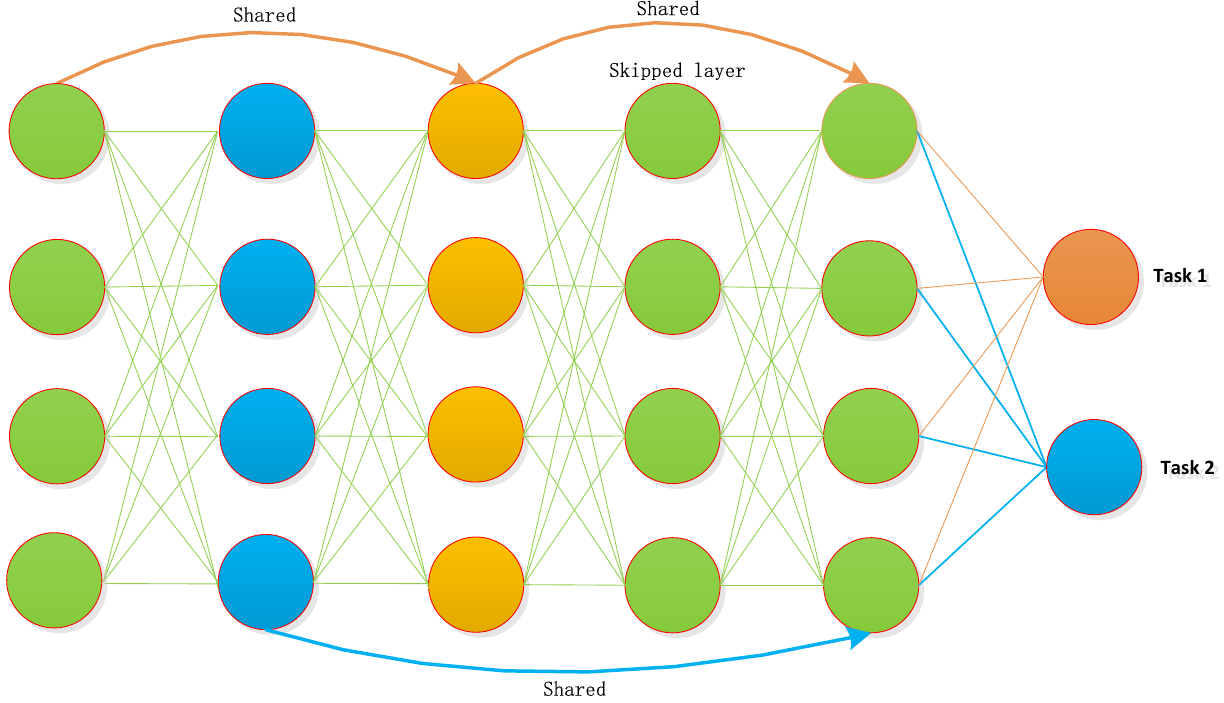}
\caption{Homogeneous MTL based on hybrid sharing.} \label{Fig:dd}
\label{fig1}
\end{figure}

3) Hybrid sharing MTL. As shown in Fig.\eqref{Fig:dd}, in the hybrid shared architecture, specific task strategies are used to select which layer of multiple task network models can
perform shared learning. For example, \cite{Sun2019AdaShareLW} uses task-specific strategies to learn shared patterns for image semantic and normal segmentation. The advantage of this method is that the number of parameters does not increase as the number of tasks does. The disadvantage is that it cannot handle heterogeneous tasks.

\subsection{Heterogeneous MTL}
HMTL usually assumes that the input (feature) space is the same, but the output (label) space is unnecessarily consistent, or the input (feature) space is heterogeneous and the output (label) space is consistent. Such methods share heterogeneous features to make some predictions such as the heterogeneous attributes of human faces and human poses, and classify various images or texts. For these existing works, we can also divide them into two categories to the HMTL sharing architectures: hierarchical sharing and sparse sharing.

\begin{figure}[ht]
\centering
\includegraphics[scale=0.53]{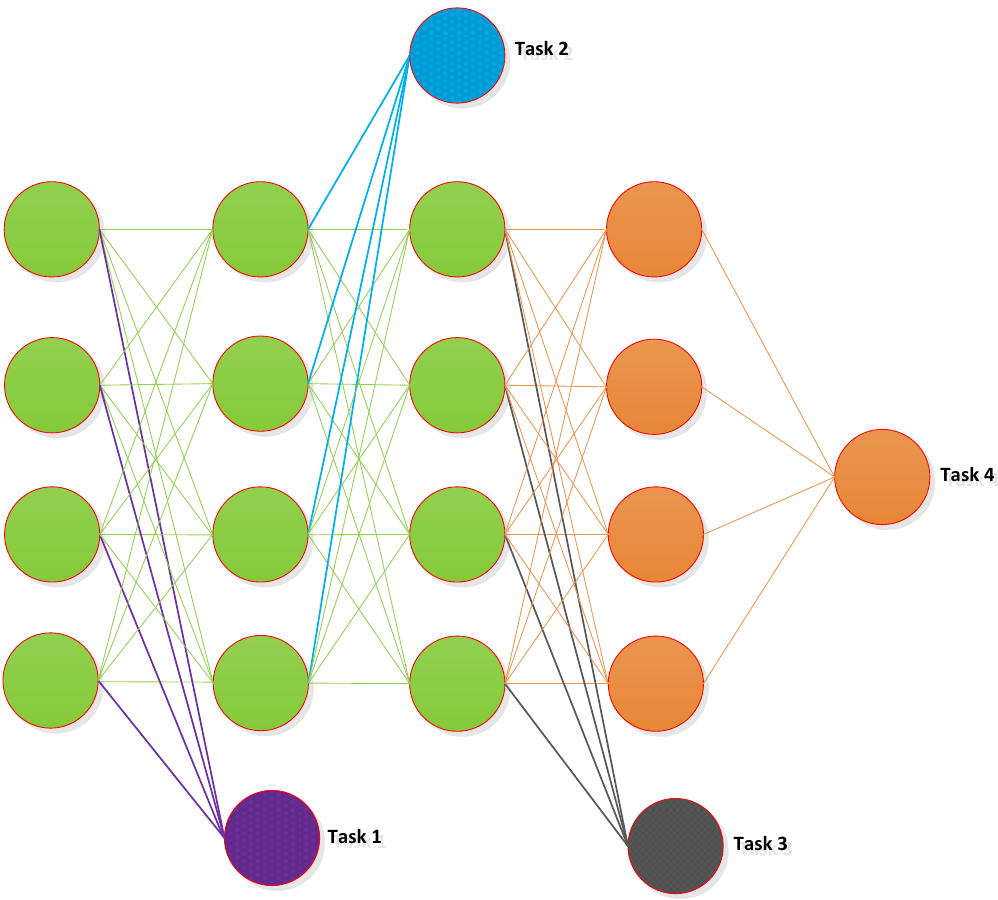}
\caption{HMTL based on hierarchical sharing.}\label{Fig:ee}
\label{fig1}
\end{figure}

1) Hierarchical sharing HMTL. As shown in Fig.\eqref{Fig:ee}, in the architecture, various tasks perform hierarchical sharing learning in multiple task networks. For example, \cite{cai2013heterogeneous} is used for image classification by weighting and sharing similar features among different levels in the multi-task network. \cite{Sanh2019AHM} learns a set of shared semantic representations from the bottom of the supervised model to multiple task hierarchies at the top of the model for natural language processing.

\begin{figure}[ht]
\centering
\includegraphics[scale=0.58]{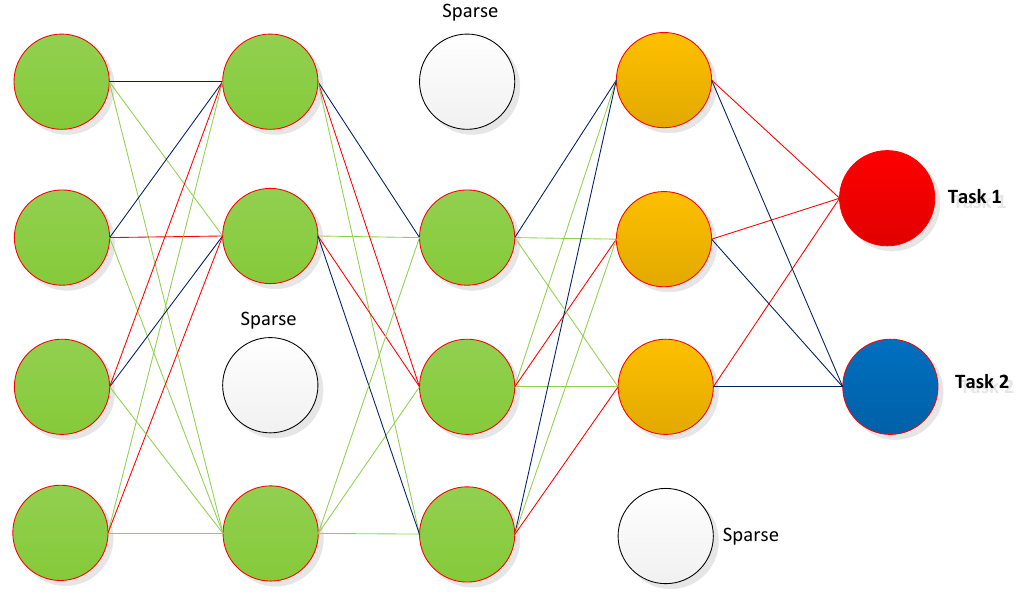}
\caption{HMTL based on sparse sharing.} \label{Fig:ff}
\label{fig1}
\end{figure}

2) Sparse sharing HMTL. As shown in Fig.\eqref{Fig:ff}, in the sparse sharing architecture, an HMT network is composed of various task networks and is sparse. For example, \cite{sun2020learning} extracts sub-networks of different tasks from an over-parameterized base network and uses masks to sparsify the features of different individual task networks to retain partly shareable features and delete irrelevant features for the Part-of-Speech , Named Entity Recognition and Chunking. The advantage of this method is to retain some useful information to help a specific task, and remove the useless information. Obviously, sparse sharing from the perspective of information sharing can be regarded as an example of hard sharing and hierarchical sharing.

On the other hand, there are also typical works not using deep learning architecture in dealing with HMTL problems. For example, \cite{zhang2011multi} uses a linear discriminant analysis of multi-task expansion algorithm (MTDA) for processing tasks with different data representations. The method learns a separate heterogeneous feature transformation for each task. It's purpose is to alleviate the problem of insufficient label data during learning, and can jointly handle binary and multi-class problems for each task. \cite{zhuang2017semantic} uses non-negative matrix factorization to learn sharing common semantic features in the feature space across heterogeneous tasks for image classification. \cite{cai2011heterogeneous} learns a shared graph Laplacian matrix in unified image features for visual clustering of different modalities.

\section{Our Method}
\begin{figure}[ht]
\centering
\includegraphics[scale=0.5]{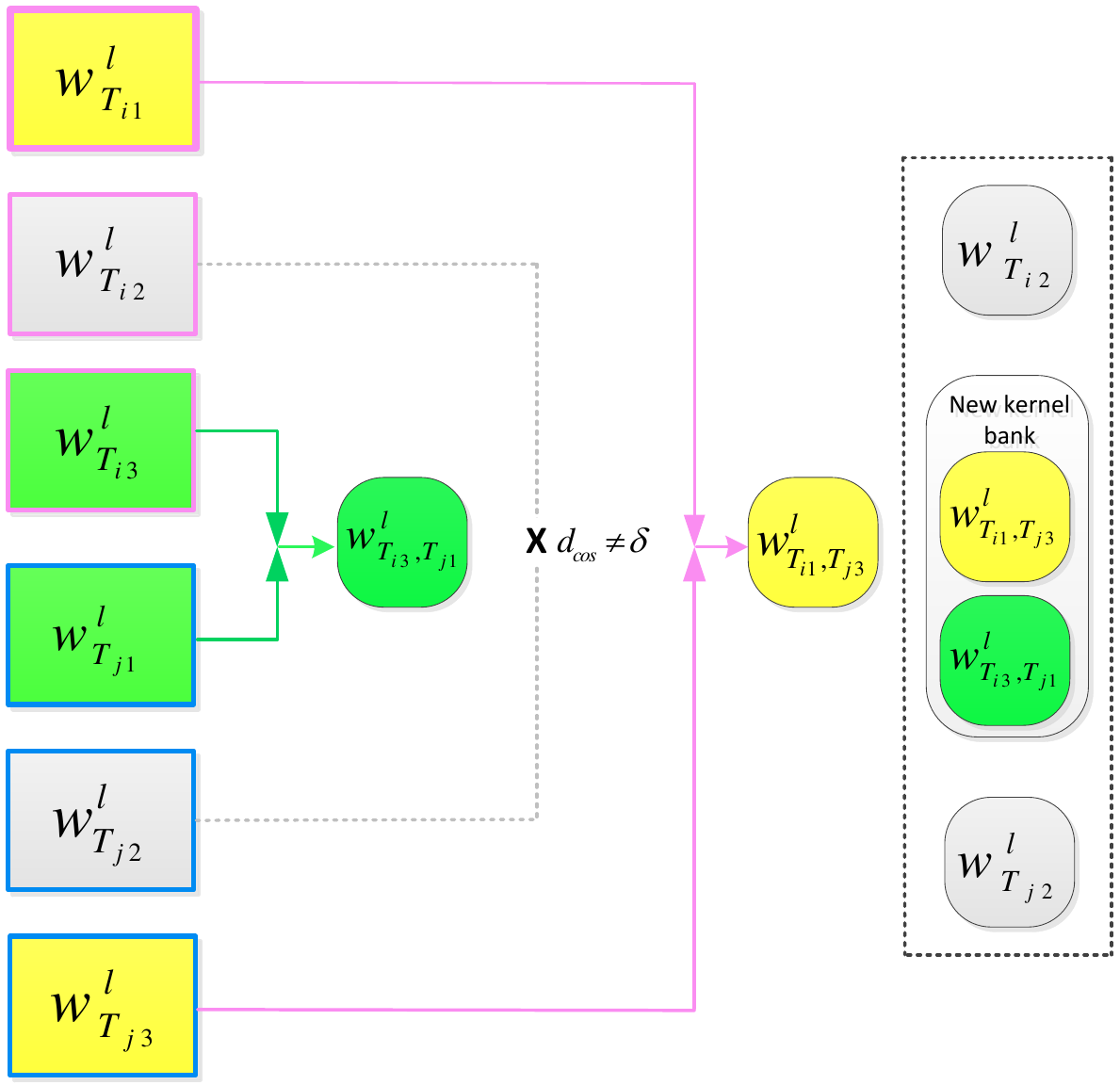}
\caption{Convolution kernel pairs sharing mechanism. In the $l$-th layer of the MTAL network, the model measures the similarity of convolution kernels in task networks $T_1$ and $T_2$ while selects the suitable convolution kernel pairs through the threshold for aggregation to form a set of new kernel bank. The purple and blue rectangles represent the convolution kernels of task networks $T_1$ and $T_2$ , respectively. The yellow and cyan octagons represent the new kernels aggregated. The solid line box on the right represents the new kernel bank. The gray octagons represent independent kernels that are not shared to continually learning their corresponding tasks.}\label{Fig:gg}
\label{fig1}
\end{figure}

Our method is inspired by the fact that the activation maps generated by similar convolution kernels in each convolution layer are also similar \cite{he2019meta} \cite{slizovskaia2019case}. For this reason, our method aims to automatically select similar convolution kernel pairs across-task to obtain common knowledge for THMTL. As shown in Fig.\eqref{Fig:gg}, our method consists of measuring the similarity of convolution kernels, selecting suitable kernel pairs, and aggregating them to form a set of new kernel bank. Specifically, Section 3.1 formally formulates the problem. Section 3.2 illustrates how to find similar convolutional kernels in the convolutional layer of the network and utilize a soft threshold to select suitable kernel pairs. How to aggregate these kernel pairs is discussed in Section 3.3. Section 3.4 derives the objective function of THMTL to deploy such a mechanism.

\subsection{Problem formulation}
Suppose we are given $N$ tasks $\left\{T_{i}\right\}_{i=1}^{N}$, some of which are related or unrelated. The training dataset $D_{i}=\left\{\left(x_{h}^{i}, y_{h}^{i}\right)\right\}_{h=1}^{n_{i}}$ for ${T_{i}}$ contains ${n_{i}}$ samples with $x_{h}^{i} \in \mathbb{R}^{d_{i}}$ and its corresponding label $y_{h}^{i} \in\left\{1, \ldots, c_{i}\right\}$, where ${d_{i}}$ and ${c_{i}}$ are the numbers of dimensions and classes in the dataset ${D_{i}}$, respectively. We do not assume that the datasets from different tasks share the same feature space, so the dimensions of feature spaces can be different. This makes MTAL applicable under more general settings than most existing handling heterogeneous task methods.

\subsection{Measure the similarity of the convolution kernels}
The goal of HMTL is to improve the generalization performance of individual tasks by sharing their relevant information. However, such information is generally unavailable and implicit. To mine this information, there have been some works proposed, they can mainly be summarized as HMTLs based on shallow and deep networks according to the architecture. Here we only focus on the deep HMTL under the same network structures since the deep networks has been more satisfactorily applied in the MTL fileds, which can be subdivide into two types according to sharing way:

1) \emph{feature}-sharing-based. This type of method is to learn a common representation from different task features in the same hypothetical space, thereby effectively helping to learn each task. Typically, \cite{lin2016multi} uses tensors to represent the feature interactions from different tasks in the same shared subspace for inductive transfer of related information, thereby providing better generalization performance for multi-task models. \cite{zhang2019multi} uses the $\ell_{\{1,2\}}$ norm to regularize the weight matrix to extract relevant features between tasks for learning multi-tasks with different feature dimensions. These methods will lower their generalization performance when the tasks are unrelated or the distributions of the data are different.

2) \emph{parameter/weight}-sharing-based. This type of methods mainly learns multiple tasks jointly by sharing common parameters hidden in the weights of different task models. According to different implementation manners, we further divide it into the following three sub-types: \emph{1)} sharing the parameters between task models based on the same space assumption: for example, \cite{8170321} shares weight parameters in different tasks in the same subspace for face detection, key point positioning, pose estimation, and gender prediction. \emph{2)} parameter matrix factorization based: e.g., \cite{2020Predicting} uses the matrix tri-factorization to process collective matrices associated with different tasks and performs joint learning to predict two types of drug-disease associations. \emph{3)} equal prior shared assumptions based: e.g., \cite{zheng2019metadata} uses a kind of meta data (i.e., contextual attributes) as a priori information to capture the relationships between different tasks for multiple tasks clustering. The above methods mainly utilize model parameters to associate different tasks. However, it is a great challenge to design an appropriate implementation manner in different tasks to obtain shared parameters. To demonstrate the importance of these parameter sharing manners in HMTL, we further studied the current two excellent deep network frameworks based on parameter sharing that are typically capable of handling heterogeneous tasks. E.g., the cross stitch network \cite{misra2016cross} connects the low-level layers of different single-task networks by learning the  parameter $\alpha$ in the task features, which is defined as

\begin{equation}
\left[\begin{array}{c}\tilde{x}_{A}^{i j} \\ \tilde{x}_{B}^{i j}\end{array}\right]=\left[\begin{array}{cc}\alpha_{A A} & \alpha_{A B} \\ \alpha_{B A} & \alpha_{B B}\end{array}\right]\left[\begin{array}{c}x_{A}^{i j} \\ x_{B}^{i j}\end{array}\right]
\end{equation}
where ${x}_{A}^{i j}$ and ${x}_{B}^{i j}$ represent the activation-maps learned in the two sub-networks, and are linearly combined to realize the information interaction in the neurons of the hidden layer, thereby outputting new hidden features $\tilde{x}_{A}^{i j}$ and $\tilde{x}_{B}^{i j}$. The SubNetwork Routing (SNR) \cite{ma2019snr} decomposes the shared-bottom module of the MMOE \cite{ma2018modeling} network into three subnetworks and shares the parameters $z$ to learn relevant information among different tasks, which is defined as
\begin{equation}
\left[\begin{array}{l}v_{1} \\ v_{2}\end{array}\right]=\left[\begin{array}{lll}z_{11} W_{11} & z_{12} W_{12} & z_{13} W_{13} \\ z_{21} W_{21} & z_{22} W_{22} & z_{23} W_{23}\end{array}\right]\left[\begin{array}{l}u_{1} \\ u_{2} \\ u_{3}\end{array}\right]
\end{equation}
where $u_{1}$, $u_{2}$ and $u_{3}$ represent the outputs by the hidden layer of each lower-level sub-networks, $W$ is the transformation matrix, $v_{1}$ and $v_{2}$ represent the inputs of the higher-lever sub-networks of the next layer. Instead of the above methods of sharing neuron units in the hidden layers of the network, our method is to share the set of neurons in (similar) convolution kernel pairs to improve the generalization and efficiency of the MTAL network.

In order to implement the above sharing, we assume that in the $l$-th layer for an individual task, there is a group of $m$ convolution kernels ${\rm W}_{T_{i}}^{l}=\left \lbrace {w}_{{T_{i}}_1}^{l},{w}_{{T_{i}}_2}^{l} \ldots, {w}_{{T_{i}}_m}^{l}\right\rbrace$, $(i=1,2,...,N)$, which form  $\mathbf{W}^{l}=\left \lbrace {\rm W}_{T_{1}}^{l},\right.$ $\left.{\rm W}_{T_{2}}^{l},\dots, {\rm W}_{T_{N}}^{l}\right\rbrace$ in the MTAL network. Then, we measure the similarity between convolution kernel pairs by the following cosine similarity to capture the relationship between tasks:
\begin{equation}\label{Eq:1}
d_ {cos} ({vec}(w_{T_i}^{l}),{vec}(w_{T_j}^{l}))= \frac {\operatorname{\emph{vec}}(w_{T_{i}}^{l})^{\mathrm{T}} \cdot \operatorname{\emph{vec}}(w_{T_{j}}^{l})} {{\|\|{\emph{vec}}(w_{T_{i}}^{l})\|\|}_{2} \cdot {\|\| {\emph{vec}}(w_{T_{j}}^{l})\|\|}_{2}}
\end{equation}
where ${vec}{(\cdot)}$ represents the vectorization operator.

Next, we just consider sharing the kernel pairs that satisfy formulation \eqref{eq:aa} while retaining the remaining independent kernels to prevent possible negative transfer caused by existing MTAL network learning.
\begin{equation} \label{eq:aa}
d_ {cos} ({vec}(w_{T_i}^{l}),{vec}(w_{T_j}^{l}))\geq \delta, \quad \quad\quad\quad\delta \in[0.1,0.9]
\end{equation}

\subsection{Aggregation of Convolution Kernels}
To further perform the sharing, we use the weighted aggregation of pairwise $w_{T_{i}}^{l}$ and $w_{T_{j}}^{l}$ to model the shared representation between individual tasks as follows:
\begin{equation} \label{eq:sharing}
\left\{{w_{T_{i}, T_{j}}^{l}} \left|\right. {w_{T_{i},T_{j}}^{l}}=\varphi_{i, j}^{l} w_{T_{i}}^{l}+\varphi_{j, i}^{l} w_{T_{j}}^{l}, \quad i\neq j, \quad\varphi_{j, i}^{l}+\varphi_{j, i}^{l}=1\right\}
\end{equation}
where $\varphi_{i, j}^{l}$ and $\varphi_{j, i}^{l}$ are weight coefficients, ${w_{T_{i} T_{j}}^{l}}$ represents the aggregated kernels. In this way, we reduce the number of weights in the entire network while suppressing intra-redundancy (due to over-parameterization \cite{denil2013predicting}). Our experiments show that the memory of the entire network can be saved by about $13.1\%$ compared to the original structures, as empirically analyzed in \ref{sec:SFIMDL}.

For joint learning of multiple tasks in the MTAL network, we gather \eqref{eq:sharing} to form a kernel bank and perform a simple average aggregation as follows:
\begin{equation}\label{Eq:6}
\widehat{w}_{T_{i}}^{l}=\frac{\sum_{w \in \mathbf{M}_{T_{i}}^{l}} w_{\mathbf{M}_{T_{i}}}^{l}}{\left|\mathbf{M}_{T_{i}}^{l}\right|}
\end{equation}
where $\widehat{w}_{T_{i}}^{l}$ represents the averaged kernel, $\mathbf{M}_{T_{i}}^{l}$  is the new kernel bank (i.e., $\mathbf{M}_{T_{i}}^{l}=\left\{w_{T_{i}, T_{j}}^{l}, \ldots, w_{T_{i}, T_{N}}^{l}\right\}, (j=2, \ldots, N, j \neq i)$), $w_{\mathbf{M}_{T_{i}}}^{l}$ is the kernel of the bank.

\subsection{Objective Function of individual $\&$ total task}
In the MTAL learning, we use the inputs $X_{T_{i}} $ (${i}=1,2, \ldots,N$) and the labels $Y_{T_{i}}$ (${i}=1,2,\ldots,N$) to minimize following the individual task loss function $\mathcal{L}_{\textit T_{i}}$ as follow
\begin{equation}
\mathcal{L}_{\textit T_{i}}=-\sum_{i=1}^{n_{{i}}} y_{h}(\log y^{\prime}_h)+\lambda \lVert \rm \textbf W_{\textit T_{\textit {i}}}\lVert_{F}^{2}
\end{equation}
where the first term is the cross-entropy loss of individual tasks, the second term is the $\ell_{2}$ norm regularization. $\lambda$ is an adjustment hyper-parameter. Then, we define the total objective function of the entire network as
\begin{equation}
\mathcal{L}_{\textit T_{total}}=\mathcal{L}_{\textit T_{1}}+\mathcal{L}_{\textit T_{2}}+...+\mathcal{L}_{\textit T_{N}}
\end{equation}

The whole process of the proposed method to solve THMT is summarized in Algorithm \eqref{AL:1}.

\begin{algorithm}
        \caption{MTAL} \label{AL:1}
        \label{alg:opthyperpar}
          \Notation{$T_{1}, T_{2}, \ldots, T_{N}$ denote input heterogeneous tasks;\\
          \noindent $\mathbf{W}^{l}=\left \lbrace {\rm W}_{T_{1}}^{l}, {\rm W}_{T_{2}}^{l},\dots, {\rm W}_{T_{N}}^{l}\right\rbrace$ denotes the convolution kernels of the $l$-th layer; $\eta$ denotes learning rate; $\delta$ is the threshold; $\theta$ are the weights of the entire network; $\mathbf{M}_{T_{i}}^{l}$ denotes the new kernel bank.}
          \KwIn{ $T_{1}, T_{2}, \ldots, T_{N}$, $\eta$, $\delta$}
          \KwOut{$\theta$}
                    random initialization $\mathbf{W}^{l}$\\

             \Repeat{convergence}{

                \For{all $\left\{\left(x_{h}, y_{h}\right) \in D_{T_{j}}\right\}_{j=1}^{\mathrm{N}}$}
                {
                \For{each convolution layer $l$ }
                {
                \For{each $T_i$}
                {
                    $\mathbf{M}_{T_{i}}^{l}=\{\}$ \\
                    \For{each $T_{j}(j\neq i)$}
                    {
                       The convolution kernel pairs similarity is calculated by \eqref{Eq:1}\\
                       If $d_ {cos} ({vec}(w_{T_i}^{l}),{vec}(w_{T_j}^{l}))\geq \delta$ : put ${w_{T_{i},T_{j}}^{l}}$ into $\mathbf{M}_{T_{i}}^{l}$ \\
                       Convolution kernel sharing by using \eqref{eq:sharing}\\
                       end for
                    }
                         Update weight $\widehat{w}_{T_{i}}^{l}$ by \eqref{Eq:6} \\
                    end for}

             end for}
               Calculate the output of the current sample sequence $\left\{y_{h}^{\prime}\right\}_{h=1}^{N}$\\
               Update weight $\theta$ by using back propagation algorithm\\
             end for}
             }

    \end{algorithm}

\section{Experiments}
In this section, we use VGG \cite{simonyan2014very} network as a base-model but also other networks such as ResNets \cite{he2016deep}. We conduct two sets of experiments on eight public datasets to verify the performance, that is, one set of experiments uses the prior relationships between tasks (e.g., obtained through cross-validation experiments) while the other set does not.
\subsection{Datasets}
We use the following datasets for experiments, and divide $70\%$ for training, and the remaining $30\%$ for testing, which is detailed in Table \eqref{lig:0a}.

The \emph{Office-Caltech} dataset\footnote{\url{https://people.eecs.berkeley.edu/~jhoffman/domainadapt/}}
contains the Office-Caltech10 dataset and the Office-Caltech31 dataset, each of which has a total of 2533 samples and is composed of a subset of image datasets from three different databases: Caltech, \emph{Amazon}, and Webcam. We select a set of \emph{Amazon} from the Office-Caltech31 dataset, and randomly select 10 categories, with a minimum image size of $200*150$ and a maximum image size of $900*557$.

The \emph{Office Home} dataset\footnote{\url{http://hemanthdv.org/OfficeHome-Dataset/}}
is composed of subsets of image datasets from different fields of \emph{Art}, Clipart, Product, Real-World. Each subset has 65 different categories and 15,500 images. We randomly select 10 classes in the \emph{Art} subset of the \emph{Office Home} with the image size of $117*85$ and $4384*2686$.

The \emph{Coil-20} dataset\footnote{\url{https://www.cs.columbia.edu/CAVE/software/softlib/coil-20.php}}
is a 20-object grayscale image dataset, consisting of a set of 720 unprocessed images of 10 objects and another set of 1,440 normalized image datasets
of 20 objects. The image acquisition comes from placing the object on the electric turntable against a black background, rotate the turntable by 360 degrees to capture the pose of the object with a fixed camera or take an image of the object by rotating the turntable by 5 degrees. We select the first set of unprocessed image datasets for the experiment.

The \emph{Chars74K} dataset\footnote{\url{http://www.ee.surrey.ac.uk/CVSSP/demos/chars74k/}}
is composed of two datasets of English characters and Kannada characters, where the English characters include 3 datasets of \emph{A} (A-Z), \emph{a} (a-z), and 0-9 handwritten digits (\emph{HD}) with a total of 62 categories, 3410 images, handwritten by 55 volunteers. We use the \emph{A}, \emph{a}, \emph{HD} subsets in the English character datasets, and randomly select 10 categories from the \emph{A}, \emph{a} subset for the experiment.

The \emph{Typographic} dataset\footnote{\url{http://www.catalina.com.cn/info_249972.html}}
is a classic dataset used for machine learning, image classification, and image recognition. It is mainly composed of 0-9 typographic numbers, a
total of 10,000 pictures, and the picture size is $12*16$.

\begin{table}
\caption{Parameters and settings for each datasets.}\label{lig:0a}
\begin{tabular}{cccc}
\toprule  
Datasets&Number of categories&Size of image&Dataset partition ($\%$)\\
\midrule  
\emph{Amazon}&10&$150*900$/$557*28$&$70$/$30$\\
\emph{Art}&10&$117*85$/$4384*2686$&$70$/$30$\\
\emph{Coil-20}&10&$128*128$&$70$/$30$\\
\emph{HD}&10&$28*28$&$70$/$30$\\
\emph{Typographic}&10&$12*16$&$70$/$30$\\
\emph{a}&10&$28*28$&$70$/$30$\\
\emph{A}&10&$28*28$&$70$/$30$\\
\bottomrule 
\end{tabular}
\end{table}

\subsection{Comparison Methods}
We compare our method with the following five representative methods.

Single task\footnote{\url{https://github.com/machrisaa/tensorflow-vgg}} baseline: This method uses a single VGG network to learn the predictive model for each independent task.

Multi-task\footnote{\url{https://github.com/luntai/VGG16_Keras_TensorFlow}} baseline: This method uses multiple identical VGG networks to jointly learn a multi-task prediction model.

MTDA\footnote{\url{https://yuzhanghk.github.io/}} \cite{zhang2011multi}: This method uses linear discriminant analysis to handle multiple tasks represented by different data.

Cross-Stitch network \footnote{\url{https://github.com/helloyide/Cross-stitch-Networks-for-Multi-task-Learning}} \cite{misra2016cross}: This method introduces the cross stitch unit in the convolutional neural network of two different tasks for knowledge sharing, thereby improving the learning performance of the network.

NDDR-CNN\footnote{\url{https://github.com/ethanygao/NDDR-CNN}} \cite{gao2019nddr}: This method performs feature fusion at each layer on different tasks to obtain shared information, thereby improving the predicting accuracy of the model.

MTAL\footnote{\url{http://parnec.nuaa.edu.cn/3021/list.htm}}: This is our proposed method, which mainly aims at automatically selecting similar convolution kernel pairs across tasks to obtain common information for THMTL.

$\rm MTAL_{R}$\footnote{\url{http://parnec.nuaa.edu.cn/3021/list.htm}}: This is our proposed method, which uses the prior relationship between tasks in the experiment.

\subsection{Hyper-Parameter Tuning}
In the contrasted deep neural network methods, we adjust the hidden units, learning rate, and the number of training steps in each layer according to the parameter settings of the corresponding references. In MTAL, we adjusted the hyper-parameters in the same way and chose the stochastic gradient descent method as the network optimizer. Specifically, we set the learning rate $\eta$ to 0.01 and $\lambda$ to 0.1. For $\delta$, we have verified through multiple experiments that its value is 0.4 when the tasks are related, but it is 0.55 when the tasks are unrelated. All the deep models are implemented by Tensorflow.

\subsection{Results of Model Performance}
We respectively show the performance of various methods on the datasets \emph{HD}, \emph{a}, \emph{Typographic}, \emph{A}, \emph{Coil-20}, \emph{Art}, and \emph{Amazon}. The detailed analysis is show as follows:

Firstly, when using the relationship between tasks, we find from Table\eqref{lig:0b} and Table\eqref{lig:0d}, that the accuracy of $\rm MTAL_{R}$ is better than most methods. However, when the relationship between tasks is not used, we find from Table\eqref{lig:0c} and Table\eqref{lig:0e} that the accuracy of MTAL is better than other methods.

\begin{table}[ht]
\centering
\caption{Performance comparison among various methods that use task relationships on \emph{HD}, \emph{a}, \emph{typographic}, and \emph{A} datasets. Among them, the bold numbers are the best classification results, and the underlined numbers are the second-best classification results.}\label{lig:0b}
\scriptsize
\label{my-label}
\begin{tabular}{c|c|c|c|cc}
\hline
\multirow{2}{*}{Methods} &\multicolumn{2}{c|}{Group 1}&\multicolumn{2}{c}{Group 2}\\  \cline{2-6}
\multicolumn{1}{c|}{}&\emph{HD}&\emph{a}&\emph{Typographic}&\emph{A}& \\ \hline
Single-task & $0.76\pm0.025$ & $0.80\pm0.023$ &$0.95\pm0.011$ & $0.84\pm0.020$ & \\
Multi-task & $\mathbf{0.83\pm0.015}$ & $0.85\pm0.015$ &$\underline{0.97\pm0.015}$ & $0.86\pm0.020$ & \\
MTDA & $0.78\pm0.032$ & $0.82\pm0.014$ &$0.95\pm0.015$ & $0.85\pm0.040$ &  \\
Cross-Stitch & $0.81\pm0.017$ & $0.86\pm0.012$ &$\underline{0.97\pm0.015}$ & $\underline{0.88\pm0.015}$ & \\
NDDR-CNN &$0.80\pm0.040$ & $\underline{0.88\pm0.035}$ & $\mathbf{0.97\pm0.014}$ & $0.87\pm0.033$ & \\
$\rm MTAL_{R}$&$\underline{0.83\pm0.035}$ & $\mathbf{0.92\pm0.029}$ &$\mathbf{0.97\pm0.014}$ & $\mathbf{0.98\pm0.013}$ & \\\hline
\end{tabular}
\end{table}

Secondly, in Table\eqref{lig:0b} and Table\eqref{lig:0d} we find that most of the MTL methods are better than the single-task learning method, which proves the effectiveness of jointly learning multiple heterogeneous tasks by exploring the relationship between tasks. In particular, Table\eqref{lig:0b} shows that all methods are better than single-task learning methods, which indicates that the more similar the relationship between tasks, the better the generalization performance of all methods. In addition, we observe that the results of different multi-task methods are different, which is caused by the differences among tasks.

Again, the results in Tables \eqref{lig:0b} to \eqref{lig:0e} show that most MTL methods using task relationships are better than those without task relationships. It suggests that most of the current MTL methods rely on the relationship between tasks. However, from these results, we find that MTAL is equivalent to the best method of the first set of experiments, and it improves the accuracy on \emph{HD} and \emph{Amazon} in Tables \eqref{lig:0c} and \eqref{lig:0e}. This further reflects the excellent performance of the convolution kernel pairs sharing mechanism.

\begin{table}[ht]
\centering
\caption{Performance comparison among various methods in \emph{HD}, \emph{A}, \emph{typographic}, and \emph{a} datasets without the task relationships. Among them, the bold numbers are the best classification results, and the underlined numbers are the second-best classification results.}\label{lig:0c}
\scriptsize
\label{my-label}
\begin{tabular}{c|c|c|c|cc}
\hline
\multicolumn{1}{c}{Models}&
\multicolumn{1}{|c}{\emph{HD}}&
\multicolumn{1}{|c}{\emph{A}}&
\multicolumn{1}{|c}{\emph{Typographic}}&
\multicolumn{1}{|c}{\emph{a}}
\\\hline
Single-task & ${0.76\pm0.025}$ & ${0.84\pm0.020}$ &${0.95\pm0.011}$ &$0.80\pm0.023$ &  \\
Multi-task & ${0.82\pm0.030}$ & ${0.85\pm0.090}$ &$0.92\pm0.012$ & $0.84\pm0.090$ & \\
MTDA & $0.75\pm0.047$ & $0.78\pm0.025$ &$0.90\pm0.024$ & $0.82\pm0.012$ &  \\
Cross-Stitch & $\underline{0.82\pm0.019}$ &${0.86\pm0.020}$ & $0.95\pm0.015$ & $\underline{0.87\pm0.090}$& \\
NDDR-CNN &$0.80\pm0.026$ & $\underline{0.87\pm0.017}$ &$\underline{0.96\pm0.005}$ & $0.85\pm0.023$ &  \\
MTAL& $\mathbf{0.86\pm0.038}$ & $\mathbf{0.98\pm0.014}$ &$\mathbf{0.97\pm0.016}$ & $\mathbf{0.92\pm0.025}$&  \\\hline
\end{tabular}
\end{table}

\begin{table}[ht]                                                                                                                                                           \centering
\caption{Performance comparison among various methods that use task relationships on \emph{HD}, \emph{Coil-20}, \emph{Art}, and \emph{Amazon} datasets. Among them, the bold numbers are the best classification results, and the underlined numbers are the second-best classification results.}\label{lig:0d}
\scriptsize
\label{my-label}
\begin{tabular}{c|c|c|c|cc}
\hline
\multirow{2}{*}{Methods} &\multicolumn{2}{c|}{Group 1}&\multicolumn{2}{c}{Group 2} \\ \cline{2-6}  
\multicolumn{1}{c|}{}&\emph{HD}&\emph{Coil-20}&\emph{Art}&\emph{Amazon}&\\ \hline
Single-task &$0.76\pm0.025$ &$\mathbf{1}$  &$0.59\pm0.041$ & ${0.76\pm0.032}$ & \\
Multi-task &$0.81\pm0.041$ & $ \mathbf{1}$ & ${0.57\pm0.059}$ & $ 0.77\pm0.042$  & \\
MTDA & $0.80\pm0.007$ & $\underline{0.97\pm0.018}$ &$0.59\pm0.013$ & $0.72\pm0.043$ &  \\
Cross-Stitch &$\underline{0.81\pm0.040}$ & $\mathbf{1}$ & ${0.60\pm0.055}$&$0.77\pm0.045$ & \\
NDDR-CNN &$0.79\pm0.041$ & $\mathbf{1}$ & $\underline{0.60\pm0.051}$  &$\underline{0.80\pm0.044}$ & \\
$\rm MTAL_{R}$&$\mathbf{0.86\pm0.034}$ & $\mathbf{1}$ & $\mathbf{0.67\pm0.050}$  &$\mathbf{0.80\pm0.043}$ & \\                                                                              \hline
\end{tabular}
\end{table}

\begin{figure} [h] \centering
\subfigure[] { \label{fig:a}
\includegraphics[width=0.4\columnwidth]{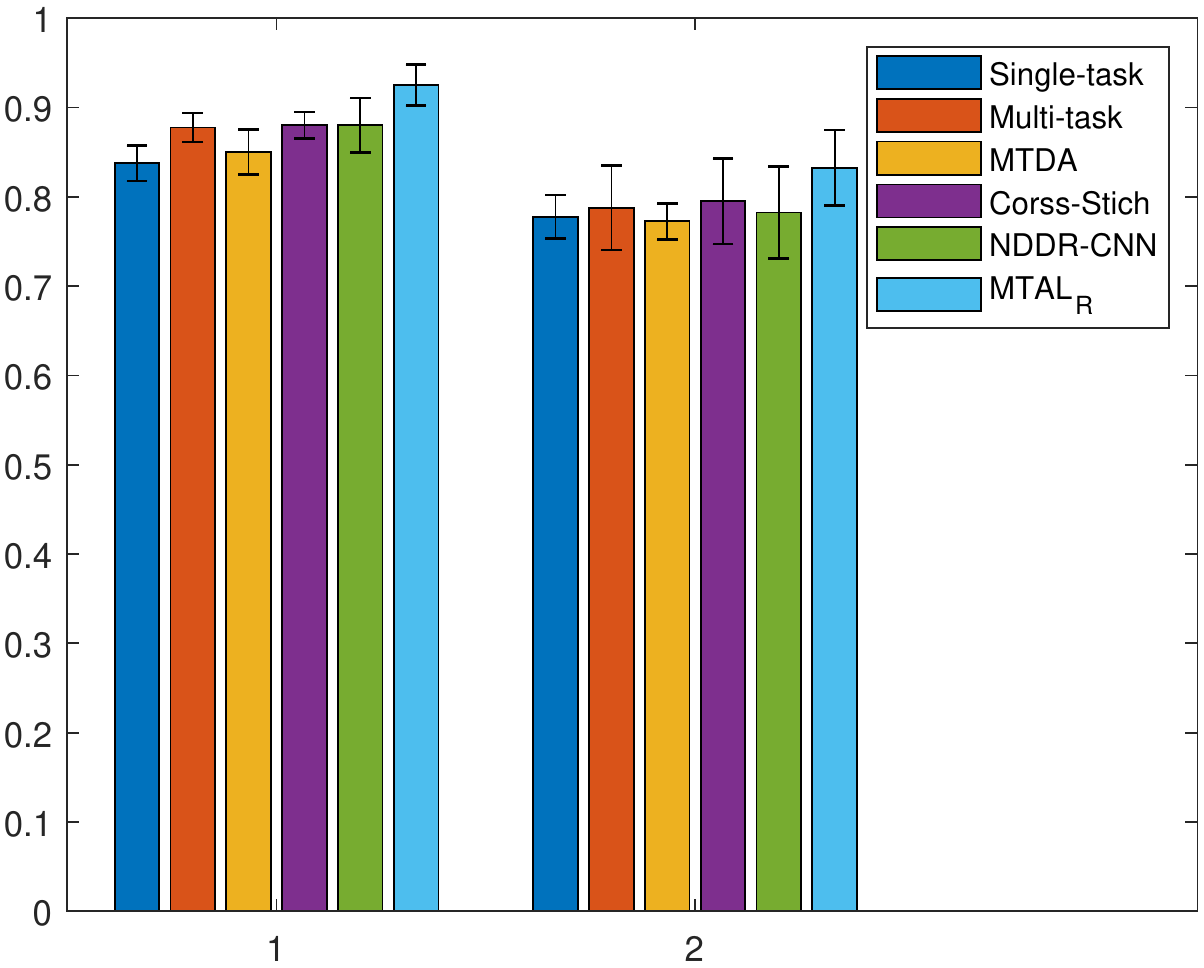}
}
\subfigure[] { \label{fig:b}
\includegraphics[width=0.4\columnwidth]{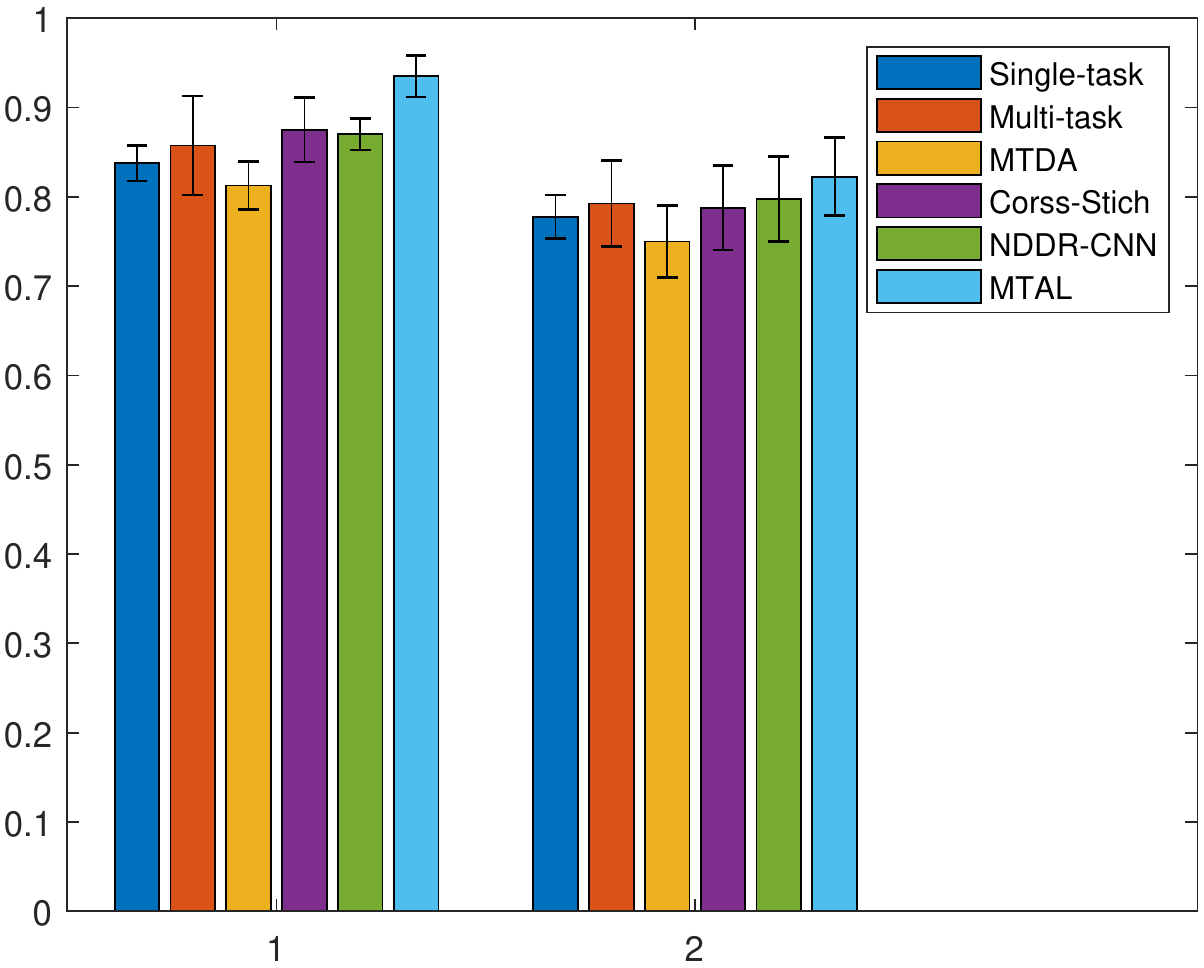}
}
\caption{Performance comparison of various methods on mean and mean square error. In Fig.8 (a), 1 and 2, the performance comparison of various methods using task relationships when the tasks are related and unrelated, respectively. In Fig.8 (b), 1 and 2, the performance comparison of various methods that do not use task relationships when tasks are related and unrelated, respectively.}\label{Fig:hh}
\label{fig}
\end{figure}

Finally, as shown in Figure \eqref{Fig:hh}, we find that the overall performance of the MTAL and $\rm MTAL_{R}$ methods is better than other methods. The above experimental results are consistent with our theoretical analysis.

\begin{table}[ht]
\centering
\caption{Performance comparison among various methods in \emph{HD}, \emph{Art}, \emph{Coil-20}, and \emph{Amazon} datasets without task relationships. Among them, the bold numbers are the best classification results, and the underlined numbers are the second-best classification results.}\label{lig:0e}
\scriptsize
\label{my-label}
\begin{tabular}{c|c|c|c|cc}
\hline
\multicolumn{1}{c}{Models}&
\multicolumn{1}{|c}{\emph{HD}}&
\multicolumn{1}{|c}{\emph{Art}}&
\multicolumn{1}{|c}{\emph{Coil-20}}&
\multicolumn{1}{|c}{\emph{Amazon}}\\
\hline
Single-task &$0.76\pm0.025$&${0.59\pm0.041}$& $\mathbf{1}$ &${0.76\pm0.032}$&\\
Multi-task &${0.79\pm0.047}$ &$\underline{0.61\pm0.060}$  &$\mathbf{1}$ &$0.77\pm0.047$  &\\
MTDA & $0.77\pm0.051$ & $0.58\pm0.025$ &$0.95\pm0.042$ & $0.70\pm0.044$ & \\
Cross-Stitch &${0.78\pm0.037}$ & $0.59\pm0.058$ &$\mathbf{1}$& $0.78\pm0.047$ & \\
NDDR-CNN &$\underline{0.78\pm0.038}$ & $0.58\pm0.063$ & $\mathbf{1}$ &$\underline{0.78\pm0.044}$ &\\
MTAL&$\mathbf{0.82\pm0.035}$  & $\mathbf{0.65\pm0.056}$ & $\mathbf{1}$ &$\mathbf{0.82\pm0.040}$ &\\\hline
\end{tabular}
\end{table}

\begin{figure} [h] \centering
\subfigure[] { \label{fig:a}
\includegraphics[width=0.458\columnwidth]{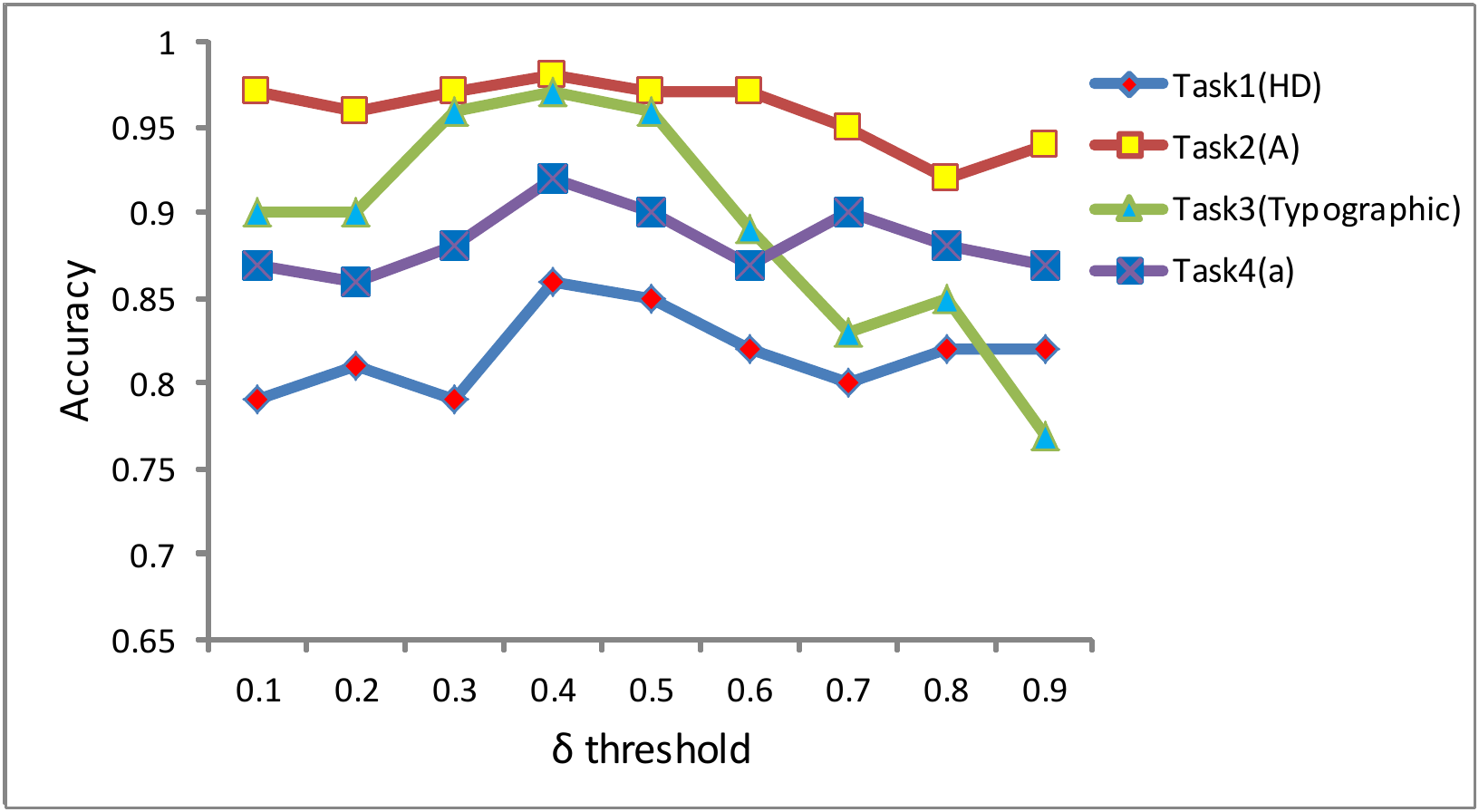}
}
\subfigure[] { \label{fig:b}
\includegraphics[width=0.462\columnwidth]{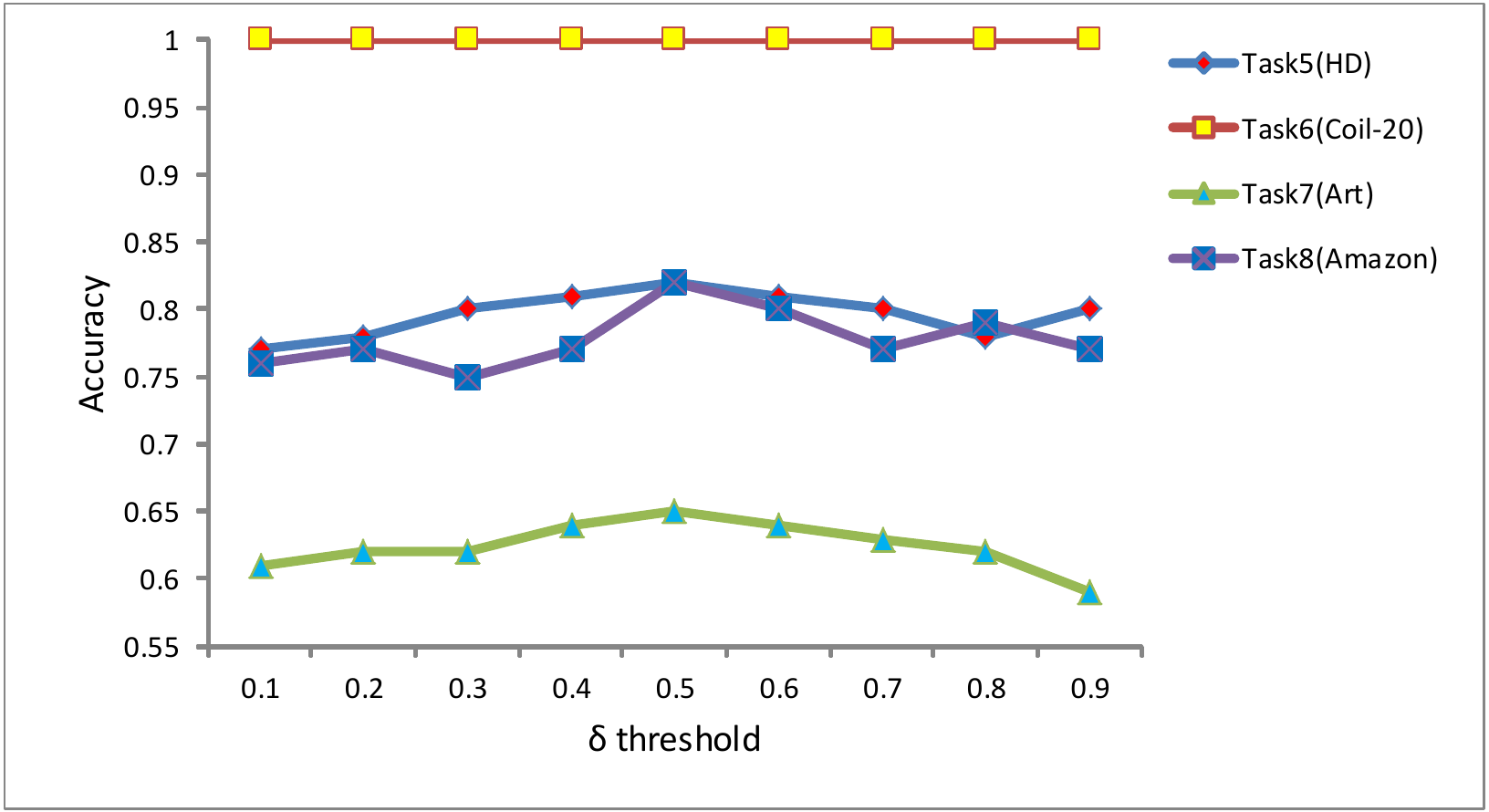}
}
\caption{The selection result of $\delta$ in various tasks. Fig.9 (a) and (b) show the corresponding learning value $\delta$ when the tasks are related and unrelated.} \label{Fig:mm}
\label{fig}
\end{figure}

\begin{table}[ht]
\caption{Kernels sharing ratio}
\begin{center}
\begin{tabular}{c|c p{0.1cm}}
\hline
MTAL network & Kernels sharing ratio ($\%$) \\ \hline
Conv1\_1 layer & $12.5$ \\ \hline
Conv1\_2 layer & $12.5$ \\\hline
Conv2\_1 layer & $16.7$ \\\hline
Conv2\_2 layer & $12.5$ \\\hline
Conv3\_1 layer & $8.3$ \\\hline
Conv3\_2 layer & $8.3$ \\\hline
Conv3\_3 layer & $12.5$ \\\hline
Conv4\_1 layer & $16.7$ \\\hline
Conv4\_2 layer & $12.5$ \\\hline
Conv4\_3 layer & $16.7$ \\\hline
Conv5\_1 layer & $12.5$ \\\hline
Conv5\_2 layer & $16.7$ \\\hline
Conv5\_3 layer & $16.7$ \\\hline
Total network & $13.1$ \\\hline
\end{tabular}\label{lig:009}
\end{center}
\end{table}

\subsection{Threshold selection analysis}
To prevent the negative transfer caused by existing MTAL network learning, we conduct threshold experiments on related and unrelated scenarios between tasks. We set the threshold in the range of 0.1-0.9, train 10 epochs for each value, and finally compute  the mean and standard deviation of classification accuracy. As shown in Fig.\eqref{Fig:mm}, we can obtain: 1) When the tasks are related and $\delta \geq 0.4$, the generalization performance of the MTAL network is the best. 2) When the tasks are unrelated and $\delta \geq 0.55$, the generalization performance of the MTAL network is the best.

\subsection{Model parameters compression and sharing strategy visualization}\label{sec:SFIMDL}
We have obtained the sharing rate (i.e., network redundancy compression) of the convolution kernels in the MTAL network and each convolution layer by experiments. As shown in Table\eqref{lig:009},  the kernel pairs shared ratios of each convolutional layer in the MTAL network are $12.5\%$, $12.5\%$, $16.7\%$, $12.5\%$, $8.3\%$, $8.3\%$, $12.5\%$, $16.7\%$, $12.5\%$, $16.7\%$, $12.5\%$, $16.7\%$, $16.7\%$  respectively. The solution space of the entire network can be compressed to $13.1\%$.

\begin{figure}[ht]
\centering
\includegraphics[scale=0.25]{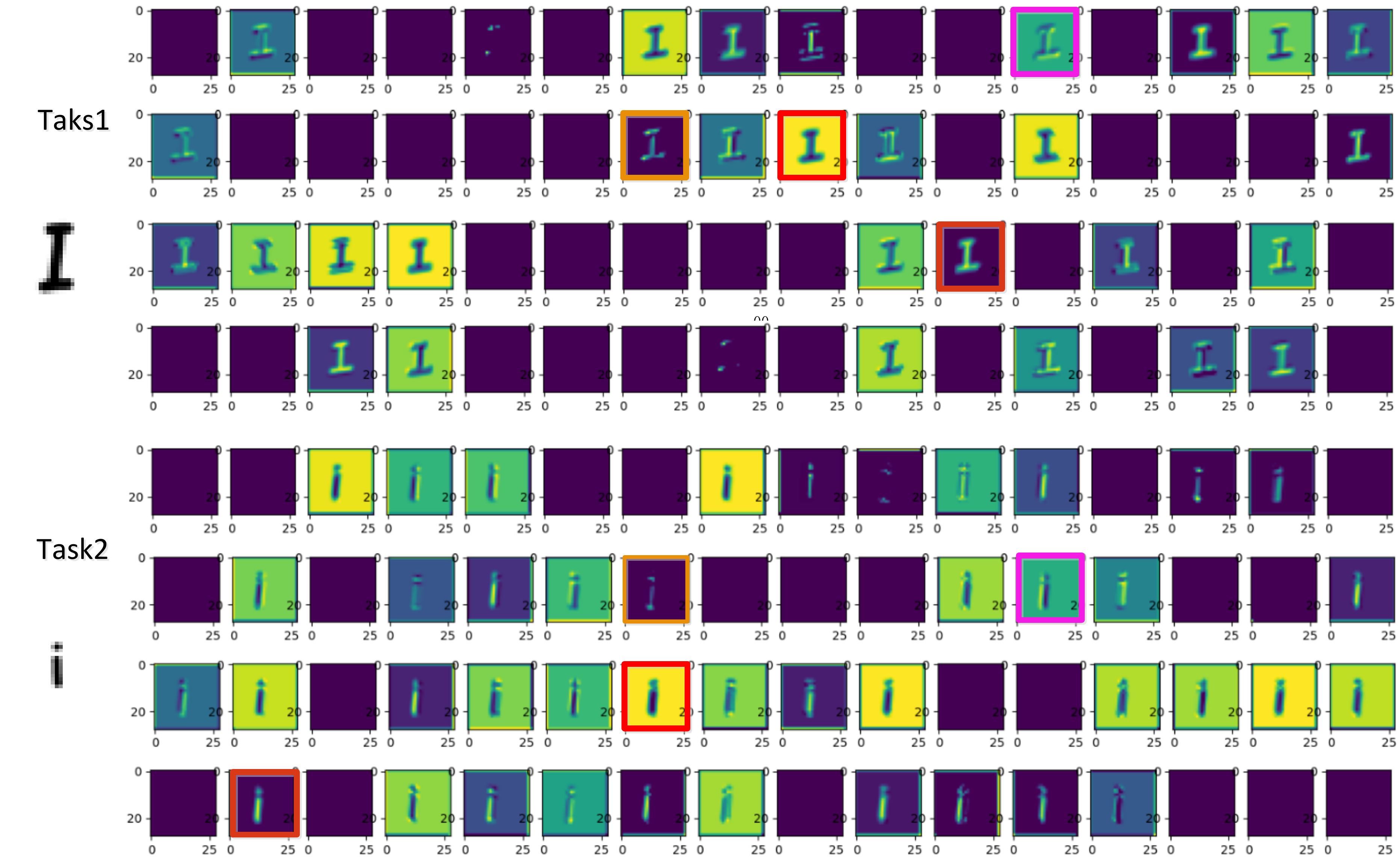}
\caption{Visual activation maps when the tasks are relevant. The activation maps are extracted from the first convolutional layer in the MTAL network. The red, purple, brown, and orange solid line boxes denote the activation maps generated from different convolution kernels in task 1 and task 2, respectively.}\label{Fig:nn}
\label{fig1}
\end{figure}

\begin{figure}[h]
\centering
\includegraphics[scale=0.25]{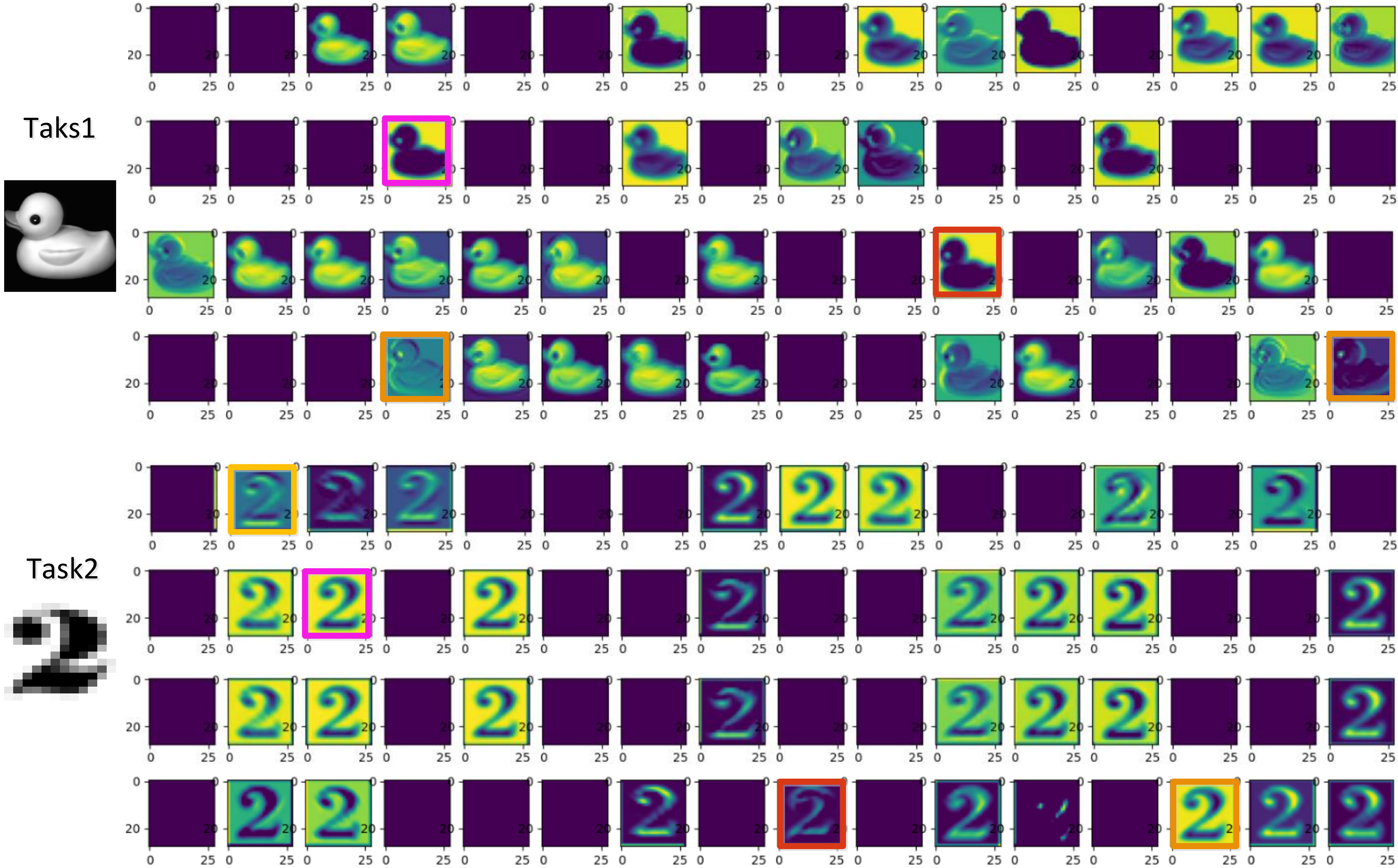}
\caption{Visual activation maps when the tasks are unrelated. The activation maps are extracted from the second convolutional layer in the MTAL network. The red, purple, brown, and orange solid line boxes denote the activation maps generated from different convolution kernels in task 1 and task 2, respectively.}\label{Fig:vv}
\label{fig1}
\end{figure}

To verify the feasibility of the convolution kernel pairs sharing mechanism proposed in this paper, we respectively visualized the activation maps in the first and second convolution layer of the MTAL network. Fig.\eqref{Fig:nn} shows the activation maps generated in the first layer of the convolution layer when the task is partially related. We find that more shapes and features information can be shared in the active maps area. Fig.\eqref{Fig:vv} shows the activation maps generate in the second convolution layer when the tasks are unrelated. We further find that more textures and contours information can be shared in the activation map area. The above visualization results show that it is feasible to perform cross-task learning by sharing similar convolution kernel pairs.

\subsection{Model convergence analysis}
\begin{figure} \centering
\subfigure[] { \label{fig:a}
\includegraphics[width=0.47\columnwidth]{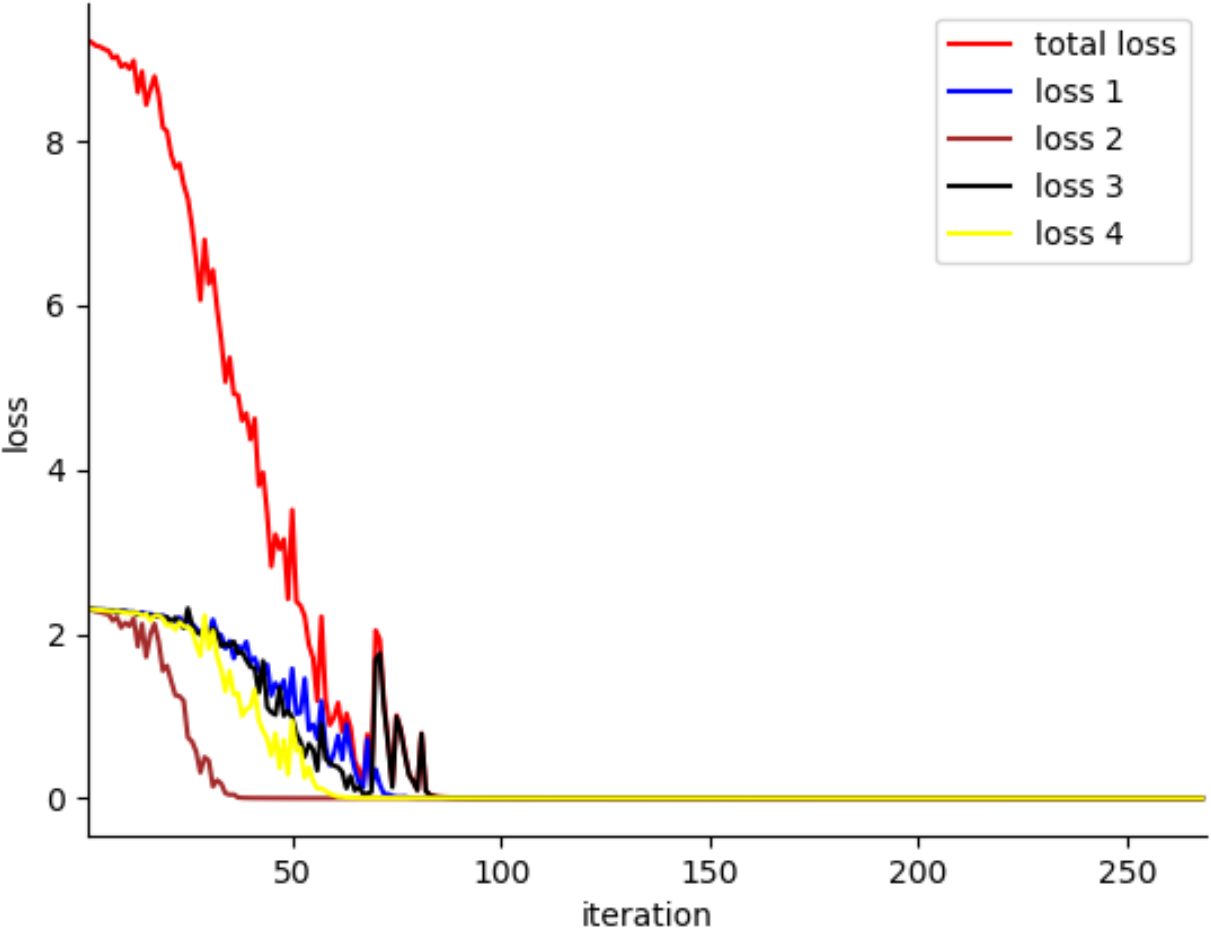}
}
\subfigure[] { \label{fig:b}
\includegraphics[width=0.47\columnwidth]{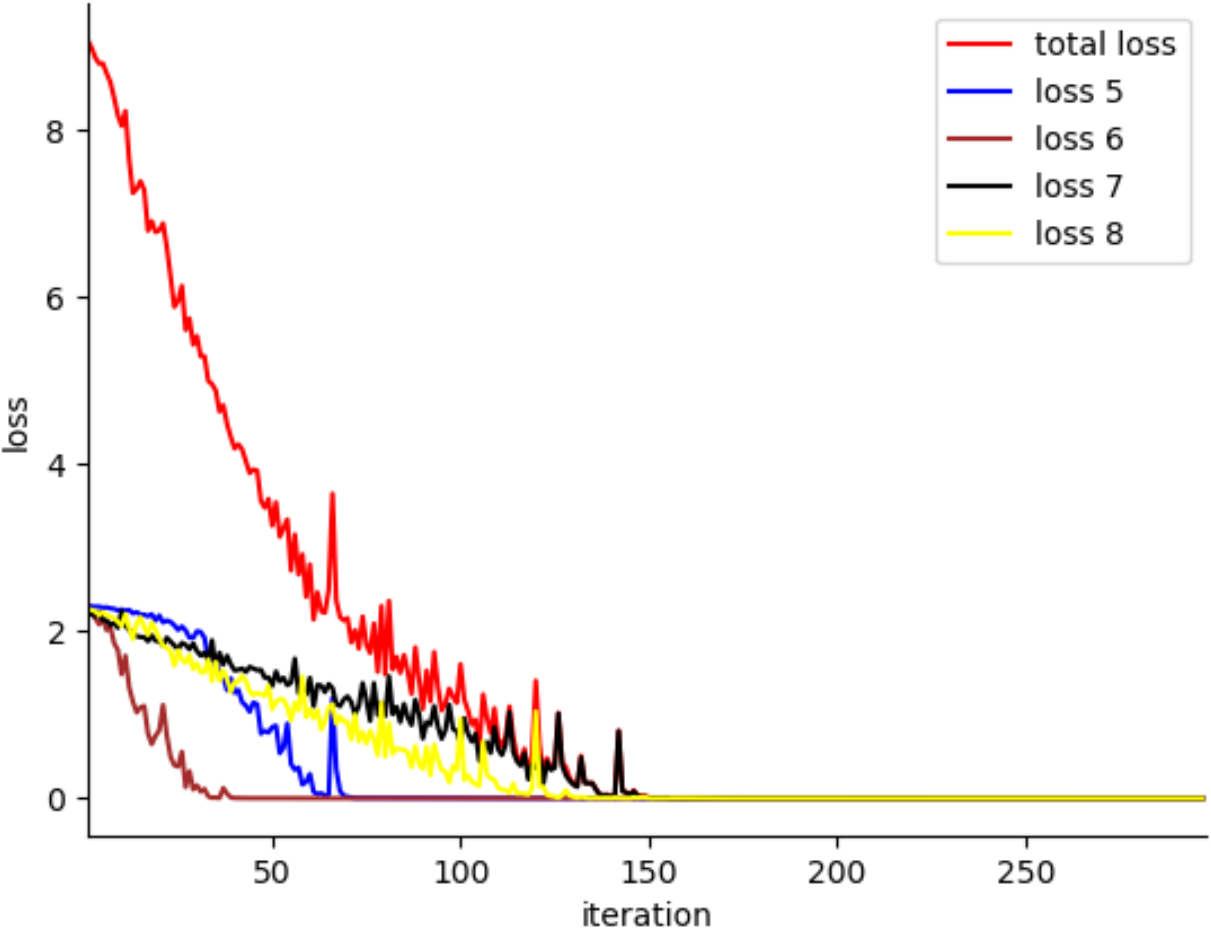}
}
\caption{Model convergence experiment. Fig. 12 (a) and (b) show the corresponding convergence curves of the model when the heterogeneous tasks are related and unrelated.}\label{Fig:uu}
\label{fig}
\end{figure}
In this section, we show the loss functions of the two sets of tasks in Fig.\eqref{Fig:uu} (a) and (b) to analyze the convergence of the model. In Figure 12(a), we find that the loss function of task 2 tends to converge after about 40 iterations, while tasks 4, 3, and 1 tend to converge after about 60, 70, and 80 iterations, respectively. In Figure 12 (b), we observe that the loss function of task 6 tends to converge after about 40 iterations, while tasks 5, 8, and 7 converge after about 80, 120, and 150 iterations, respectively. The above shows that our model converge relatively fast.
\section{Conclusion}
In this paper, we provide a deep learning framework MTAL for processing THMT. Compared with the previous MTL method, the MTAL network explores and uses the inherent relationship between tasks to share knowledge of similar convolution kernel pairs in each of their layers to learn THMT. The network not only effectively performs cross-task learning but also suppresses the intra-redundancy of the entire network. Meanwhile, MTAL can handle related heterogeneous tasks well and achieve great performance when they are unrelated. At the same time, the designed sharing strategy in MTAL can be flexibly embedded in other deep multi-task learning frameworks. To evaluate the proposed MTAL, we conduct experiments on eight public datasets and compare with the state-of-the-art HMTL methods. Experimental results show superiority of our MTAL. In summary, our work can enrich HMTL research from three aspects: 1) an adaptive THMT learning mechanism that can avoid negative transfer caused by jointly learning multiple tasks due to incorrect pre-defined task relationships; 2) a new method for solving THMT with no relation among tasks; 3) a new THMT sharing strategy for learning multiple heterogeneous tasks. However, in this work, we do not solve the problem of unbalanced and interpretable shared learning among multiple heterogeneous tasks and will devote ourselves to solving these problems next.

\section*{Acknowledgments}
This work is supported in part by Key Program of NSFC under Grant No. 61732006 and the NSFC under Grant No. 62076124.

\section*{References}

\bibliography{mybibfile}

\end{document}